\documentclass[letterpaper]{article} 
\usepackage{aaai24}  
\usepackage{times}  
\usepackage{helvet}  
\usepackage{courier}  
\usepackage[hyphens]{url}  
\usepackage{graphicx} 
\usepackage{natbib}  
\usepackage{caption} 
\usepackage{algorithm}
\usepackage{algpseudocode}

\usepackage{newfloat}
\usepackage{listings}

\DeclareCaptionStyle{ruled}{labelfont=normalfont,labelsep=colon,strut=off} 
\floatstyle{ruled}
\newfloat{listing}{tb}{lst}{}
\floatname{listing}{Listing}
\title{DGPO: Discovering Multiple Strategies with {\underline D}iversity-{\underline G}uided {\underline P}olicy {\underline O}ptimization}
\author{
    Wentse Chen\textsuperscript{\rm 1},
    Shiyu Huang\textsuperscript{\rm 2},
    Yuan Chiang\textsuperscript{\rm 3},
    Tim Pearce\textsuperscript{\rm 4},
    Wei-Wei Tu\textsuperscript{\rm 2},
    Ting Chen\textsuperscript{\rm 3},
    Jun Zhu\textsuperscript{\rm 3}
}
\affiliations{
    \textsuperscript{\rm 1}Carnegie Mellon University, Pittsburgh, USA\\
    \textsuperscript{\rm 2}4Paradigm Inc., Beijing, China\\
    \textsuperscript{\rm 3}Tsinghua University, Beijing, China\\
    \textsuperscript{\rm 4}Microsoft Research, Cambridge, United Kingdom\\
    wentsec@andrew.cmu.edu, huangsy1314@163.com, yjennice2001@gmail.com, tim.pearce@microsoft.com, tuweiwei@4paradigm.com, tingchen@tsinghua.edu.cn, dcszj@tsinghua.edu.cn

}

\usepackage{bibentry}

\usepackage{amsmath}
\usepackage{amssymb}
\usepackage{dsfont}
\usepackage{subfig}
\usepackage{booktabs}

\begin{document}

\maketitle

\begin{abstract}
Most reinforcement learning algorithms seek a single optimal strategy that solves a given task. However, it can often be valuable to learn a diverse set of solutions, for instance, to make an agent's interaction with users more engaging, or improve the robustness of a policy to an unexpected perturbance.
We propose Diversity-Guided Policy Optimization (DGPO), an on-policy algorithm that discovers multiple strategies for solving a given task. 
Unlike prior work, it achieves this with a shared policy network trained over a single run.
Specifically, we design an intrinsic reward based on an information-theoretic diversity objective. 
Our final objective alternately constraints on the diversity of the strategies and on the extrinsic reward.
We solve the constrained optimization problem by casting it as a probabilistic inference task and use policy iteration to maximize the derived lower bound.
Experimental results show that our method efficiently discovers diverse strategies in a wide variety of reinforcement learning tasks. 
Compared to baseline methods, DGPO achieves comparable rewards, while discovering more diverse strategies, and often with better sample efficiency. 
\end{abstract}

\section{Introduction}

Reinforcement Learning (RL) has pioneered breakthroughs in various domains ranging from video games~\cite{vinyals2019grandmaster,berner2019dota,huang2019combo,huang2021tikick} to robotics~\cite{Raffin18,yu2021learning}. While its achievements are remarkable, RL is not devoid of challenges. A paramount issue is the innate pursuit of RL algorithms for a singular optimal solution, even when a myriad of equally viable strategies exists. This tunnel vision for optimization can inadvertently introduce weaknesses.

For instance, RL algorithms are known for "overfitting" tasks. By zeroing in on just one strategy, they often miss out on exploring a wealth of high-quality alternative solutions. This over-specialization renders the agent vulnerable to unpredictable environmental changes, as it lacks the robustness multiple strategies could have offered \cite{kumar2020one}. In competitive arenas, predictability can be an Achilles heel, with adversaries exploiting the agent's inflexibility. A diversified approach would obfuscate the agent's strategies, improving its competitive edge \cite{lanctot2017unified}. Furthermore, in domains like dialogue systems, monotony can dull user interactions, whereas varied responses could significantly enhance user experience \cite{li2016deep,gao2019generating,pavel2020increasing,xu2022diverse,chow2022mixture}.

For instance, RL algorithms are known for "overfitting" tasks. By zeroing in on just one strategy, they often miss out on exploring a wealth of high-quality alternative solutions. This over-specialization renders the agent vulnerable to unpredictable environmental changes, as it lacks the robustness multiple strategies could have offered \cite{kumar2020one}. In competitive arenas, predictability can be an Achilles heel, with adversaries exploiting the agent's inflexibility. A diversified approach would obfuscate the agent's strategies, improving its competitive edge \cite{lanctot2017unified}. Furthermore, in domains like dialogue systems, monotony can dull user interactions, whereas varied responses could significantly enhance user experience \cite{li2016deep,gao2019generating,pavel2020increasing,xu2022diverse,chow2022mixture}.

We identify two key scenarios where multiple strategies are beneficial:
1. The margin for error is inconsequential to the task's success. In such cases, agents can operate optimally without adhering strictly to the best strategy, enabling a spread of near-optimal strategies \cite{zahavy2021discovering}.
2. The task inherently allows multiple optimal solutions, such as a maze offering two equally efficient paths \cite{osa2021discovering,zhou2022continuously}.

Crafting an algorithm that harnesses diverse high-reward solutions efficiently is intricate. With the Diversity-Guided Policy Optimization (DGPO) we propose, we aim to address several requisites for such an algorithm:

{\bf Strategy Representation}: the diversity of a policy suite is evaluated in DGPO using a metric grounded on the mutual information between states and the latent variable, implemented through a learned discriminator.

{\bf Diversity Evaluation}: The diversity of a policy suite is evaluated in DGPO using a metric grounded on the mutual information between states and the latent variable, implemented through a learned discriminator.

{\bf Diversity Exploration}: DGPO embarks on exploration that encourages deviation from familiar strategies while safeguarding performance. This is achieved using a constrained optimization method that harmonizes performance and diversity.

{\bf Sample Efficiency}: Unlike predecessors like RSPO \cite{zhou2022continuously} and RPG \cite{tang2021discovering}, which necessitated multiple networks and training phases, DGPO employs a shared network for concurrent learning of strategies, resulting in superior sample efficiency.

In encapsulation, this work makes three pivotal contributions:
(1) We introduce a structured approach to discover diverse high-reward policies by framing it as two constrained optimization problems, coupled with tailored diversity rewards to guide the policy learning.
(2) DGPO, a novel on-policy algorithm, is unveiled, designed to seamlessly uncover a diverse suite of high-quality strategies.
(3) Our empirical evaluations elucidate that DGPO not only holds its own against benchmarks but frequently surpasses them in terms of diversity, performance, and sample efficiency.


\section{Related Works}

In this section, we provide an overview of prior research that encompasses two main aspects: the representation of reinforcement learning (RL) as a probabilistic graphical model (PGM), and the explicit integration of diversity learning with RL.

\subsection{Reinforcement Learning as Probabilistic Graphical Model}
\label{sec:PGMRL}

PGM's have proven to be a useful way of framing the RL problem~\cite{ziebart2008maximum,furmston2010variational,levine2018reinforcement}. The soft actor-critic (SAC) algorithm~\cite{haarnoja2018soft} formalizes RL as probabilistic inference and maximizes an evidence lower bound by adding an entropy term to the training objective, encouraging exploration. PGM's also serve as useful tools for studying partially observable Markov decision processes (POMDP's)~\cite{igl2018deep,huang2019svqn,lee2019stochastic}. Haarnoja et al. \cite{haarnoja2018latent} use PGM's to construct a hierarchical reinforcement learning algorithm.
Hausman et al. \cite{hausman2018learning} optimize a multi-task policy through a variational bound, allowing for the discovery of multiple solutions with a minimum number of distinct skills. 
In this work, we modify the PGM of a Markov decision process (MDP) by introducing a latent variable to induce diversity into the MDP. We then derive the evidence lower bound of the new PGM, allowing us to construct a novel RL algorithm.

\subsection{Diversity in Reinforcement Learning}
Achieving diversity has been studied in various contexts in RL~\cite{mouret2009overcoming,mohamed2015variational,eysenbach2018diversity,osa2021discovering,derek2021adaptable}. Eysenbach et al. \cite{eysenbach2018diversity} proposed DIAYN to maximize the mutual information between states and skills, which results in a maximum entropy policy. Osa et al. \cite{osa2021discovering} proposed a method that can learn infinitely many solutions by training a policy conditioned on a continuous or discrete low-dimensional latent variable. Their method can learn diverse solutions in continuous control tasks via variational information maximization. There is also a growing corpus of work on diversity in multi-agent reinforcement learning~\cite{mahajan2019maven,lee2019learning,he2020skill}. 
Mahajan et al. \cite{mahajan2019maven} proposed MAVEN, a method that overcomes the detrimental effects of QMIX’s~\cite{rashid2018qmix} monotonicity constraint on exploration by maximizing the mutual information between latent variables and trajectories. However, their method can not find multiple diverse strategies for a specified task.
He et al. \cite{he2020skill} investigated multi-agent algorithms for learning diverse skills using information bottlenecks with unsupervised rewards. However, their method operates in an unsupervised manner, without external rewards. More recently, RSPO \cite{zhou2022continuously} was proposed to derive diverse strategies. However, it requires multiple training stages, which results in poor sample efficiency -- our method trains diverse strategies simultaneously which reduces sample complexity.


\section{Preliminaries}

\begin{figure}
\centering
  \includegraphics[width=0.9\linewidth]{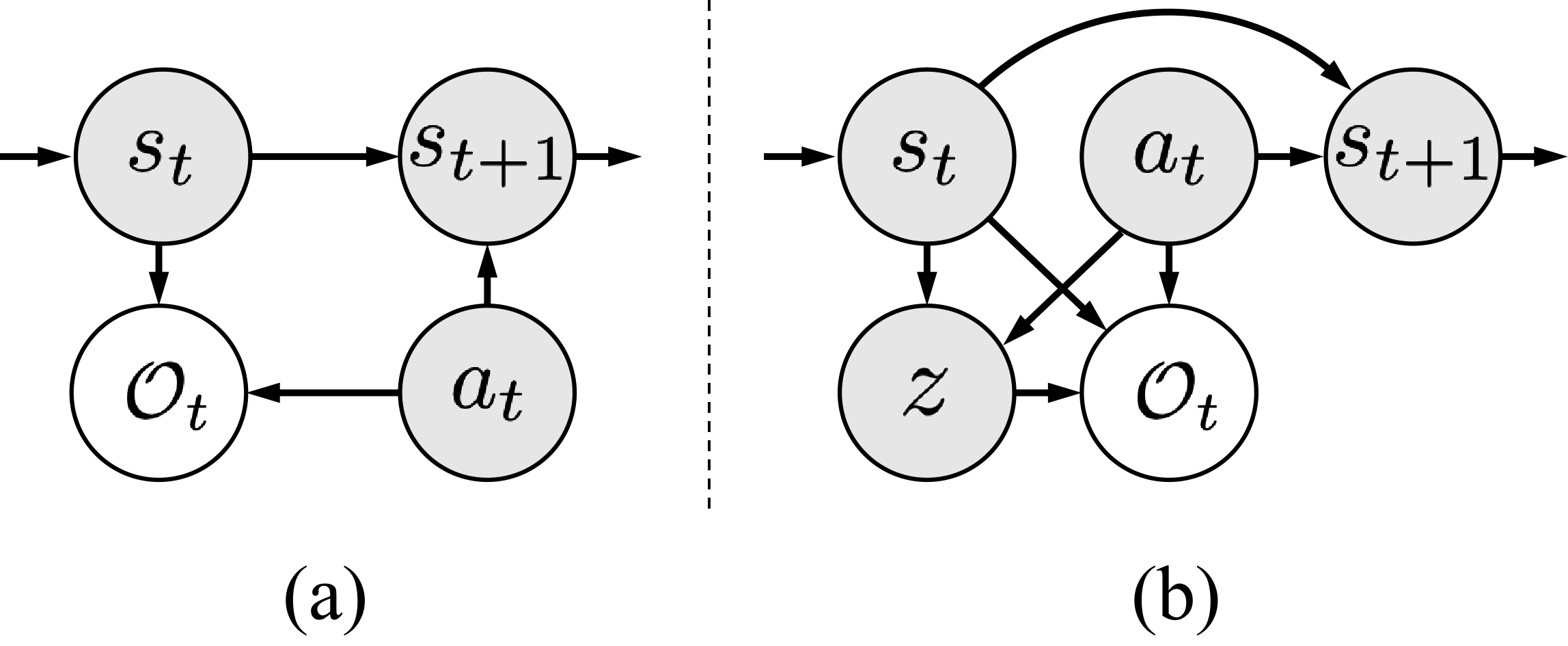}
  \caption{(a) The graphical model of MDPs. (b) The graphical model of diverse MDPs. Grey nodes are observed, and white nodes are hidden. As introduced in \cite{levine2018reinforcement}, $\mathcal{O}_t$ is a binary random variable, where $\mathcal{O}_t=1$ denotes that the action is optimal at time $t$, and $\mathcal{O}_t=0$ denotes that the action is not optimal.}
\label{fig:PGM_RL}
\end{figure}
RL can be formalized as an MDP. An MDP is a tuple $(\mathcal{S}, \mathcal{A}, P, r, \gamma)$, where $\mathcal{S}$ and $\mathcal{A}$ represent state and action space respectively, $P(s,a) : \mathcal{S} \times \mathcal{A} \rightarrow \mathcal{S}$ is the transition probability density, 
$r(s,a) :\mathcal{S} \times \mathcal{A} \rightarrow \mathbb{R}$ is a reward function, and
$\gamma\in[0,1]$ is the discount factor. 

{\bf Latent conditioned policy:} We consider a policy $\pi_\theta$ that is conditioned on latent variable $z$ to model diverse strategies, where $\theta$ represents policy parameters. For compactness, we will omit $\theta$ in our notation. We denote the latent-conditioned policy as $\pi(a|s,z)$ and a latent conditioned critic network as $V^{\pi}(s,z)$. For each episode, a single latent variable is sampled, $z\sim p(z)$ from a categorical distribution with $n_z$ categories. In our work, we choose $p(z)$ to be a uniform distribution. The agent then conditions on this latent code $z$, to produce a trajectory $\tau_{z}$.

{\bf Discounted state occupancy:} The discounted state occupancy measure for policy $\pi$ is defined as $\rho^{\pi}(s)=(1-\gamma)\sum_{t=0}^{\infty}\gamma^tP_t^{\pi}(s)$, where $P_t^{\pi}(s)$ is the probability that policy $\pi$ visit state $s$ at time $t$. The goal of the RL agent is to train a policy $\pi$ to maximize the discounted accumulated reward $J(\theta)=\mathbb{E}_{z\sim p(z),s\sim\rho^{\pi}(s), a\sim\pi(\cdot|s,z)}[\sum_t\gamma^t r(s_t,a_t)]$.

{\bf RL as probabilistic graphical model}: An MDP can be framed as a probabilistic graphical model as shown in Fig.~\ref{fig:PGM_RL}(a) and the optimal control problem can be solved as a probabilistic inference task~\cite{levine2018reinforcement}. In this paper, we propose a new probabilistic graphical model, denoted as the diverse MDP, as shown in Fig.~\ref{fig:PGM_RL}(b). We introduce a binary random variable $\mathcal{O}_t$ and an integer random variable $z$ into the model. 
$\mathcal{O}_t=1$ denotes that action $a_t$ is optimal at time step $t$ and $\mathcal{O}_t=0$ denotes it is not. 
Previous work\cite{levine2018reinforcement} has defined, $p(\mathcal{O}_t=1|s_t,a_t,z) \propto \exp(r(s_t,a_t))$. The evidence lower bound (ELBO) is given by: 
\begin{eqnarray}
\begin{aligned}
&\log p(\mathcal{O}_{1:T}) \\
&\geq \mathbb{E}_{\tau\sim \mathcal{D}_\pi}[\log p(\mathcal{O}_{1:T},a_{1:T},s_{1:T},z) \\
&\qquad-\log\pi(a_{1:T},s_{1:T},z)] \\
&= \mathbb{E}_{\tau\sim \mathcal{D}_\pi}[\log{p(\mathcal{O}_t|s_t,a_t,z)} \\
&\qquad+\log{p(z|s_t,a_t)}-\log{\pi(a_t|s_t,z)} ],\\
\end{aligned}
\label{eq:probabilistic inference}
\end{eqnarray}
where the trajectory $\tau = \{a_{1:T},s_{1:T},z\}$ is sampled from a trajectory dataset $\mathcal{D}_\pi$. The proof of Eq.~\ref{eq:probabilistic inference} can be found in Appendix C. 
Note that the introduction of $z$ gives rise to the term $p(z|s_t,a_t)$, which represents how identifiable the latent code is from the current policy behavior. This will be a crucial ingredient of DGPO, guiding the policy to explore and discover a set of diverse strategies. 


\section{Methodology}

In this section, we will introduce our algorithm in detail. Our algorithm can be divided into two stages. In the first stage, we will focus on improving the agent's performance while maintaining its diversity. In the second stage, we will focus more on enhancing diverse strategies. Finally, we will introduce our final algorithm, denoted as Diversity-Guided Policy Optimization (DGPO), that unifies the two-stage training processes and also the implementation details.

\subsection{Diversity Measurement}

In this section, we present a diversity score capable of evaluating the diversity of a given set of policies. We then derive a diversity objective from this score to facilitate exploration.
Eysenbach et al. ~\cite{eysenbach2018diversity} proposed a diversity score based on mutual information between states and latent codes $z$,

\begin{eqnarray}
\begin{aligned}
I(s;z) = \mathbb{E}_{z\sim p(z),s\sim\rho^\pi(s)}[\log p(z|s)-\log p(z)],
\end{aligned}
\label{eq:Diversity-metric-1}
\end{eqnarray}
where $I(\cdot,\cdot)$ stands for mutual information. As we can not directly calculate $p(z|s)$, we approximate it with a learned discriminator ${q}_{\phi}(z|s)$ and derive the ELBO as $\mathbb{E}_{z\sim p(z),s\sim\rho^\pi(s)}[\log q_{\phi}(z|s)-\log p(z)]$, where $\phi$ are the parameters of the discriminator network.

Mutual information is equal to the KL distance between the state marginal distribution of one policy and the average state marginal distribution, i.e., $I(s;z) = \mathbb{E}_{z\sim p(z)}[D_{KL}(\rho^\pi(s|z)||\rho^\pi(s))]$ ~\cite{eysenbach2021information}.  
This means that $I(s;z)$ captures the diversity averaged over the \textit{whole set} of policies.
In DGPO, we wish to ensure that \textit{any given pair} of strategies is different, rather than on average. As such, we define a novel, stricter diversity score,

\begin{eqnarray}
\begin{aligned}
\mathrm{DIV}(\pi_{\theta}) = \mathbb{E}_{z\sim p(z)}[\min_{z'\neq z} D_{KL}(\rho^{\pi_\theta}(s|z)||\rho^{\pi_\theta}(s|z'))].
\end{aligned}
\label{eq:Diversity-metric-2}
\end{eqnarray}

Instead of comparing policy with the average state marginal distribution, we compare it with the nearest policy in $\rho^{\pi}(s)$ space. In this way, setting $\mathrm{DIV}(\pi_\theta)\geq\delta$ means that each pair of policies have at least $\delta$ distance in terms of expectation. In order to optimize Eq.~\ref{eq:Diversity-metric-2}, we first derive a lower bound,
\begin{eqnarray}
\begin{aligned}
\mathrm{DIV}(\pi_{\theta}) \geq \mathbb{E}_{z\sim p(z),s\sim\rho^\pi(s)}\left[\min_{z'\neq z} \log\frac{p(z|s)}{p(z|s)+p(z'|s)}\right].
\end{aligned}
\label{eq: Div-lower-bound}
\end{eqnarray}

The proof is given in Appendix D. To maximize this lower bound, we first assume we can learn a latent code discriminator, ${q}_{\phi}(z|s)$, to approximate $p(z|s)$. We then define an intrinsic reward, 
\begin{eqnarray}
\begin{aligned}
r_t^{in} = \min_{z'\neq z} \log \frac{{q}_{\phi}(z|s_{t+1})}{{q}_{\phi}(z|s_{t+1})+{q}_{\phi}(z'|s_{t+1})} .
\end{aligned}
\label{eq: intrinsic-reward}
\end{eqnarray}

This allows us to define our final diversity objective,
\begin{eqnarray}
\begin{aligned}
J_{Div}(\theta) = 
\mathbb{E}_{z\sim p(z), s\sim\rho^{\pi}(s), a\sim\pi(\cdot|s,z)}[\sum_t\gamma^t r^{in}_t].
\end{aligned}
\label{eq:Diversity objective}
\end{eqnarray}

A straightforward way to incorporate the diversity metric into a PGM is defining elements in Eq.~\ref{eq:probabilistic inference} as,
\begin{eqnarray}
\begin{aligned}
& p(\mathcal{O}_t=1|s_t,a_t,z) =
\exp\left(r(s_t,a_t)\right) , \\
& p(z|s_t,a_t) = 
\exp\left(r^{in}_t\right). 
\end{aligned}
\label{eq:naive-ELBO-element}
\end{eqnarray}

However, the simple combination of extrinsic and intrinsic rewards may lead to poor performance. In the following paragraph, we formulate the algorithm as constrained optimization problems and mask elements in Eq.~\ref{eq:naive-ELBO-element} based on constraints to guide the policy to explore.

\subsection{Stage 1: Diversity-Constrained Optimization}

\begin{figure*}[t]
  \centering
  \includegraphics[width=0.93\linewidth]{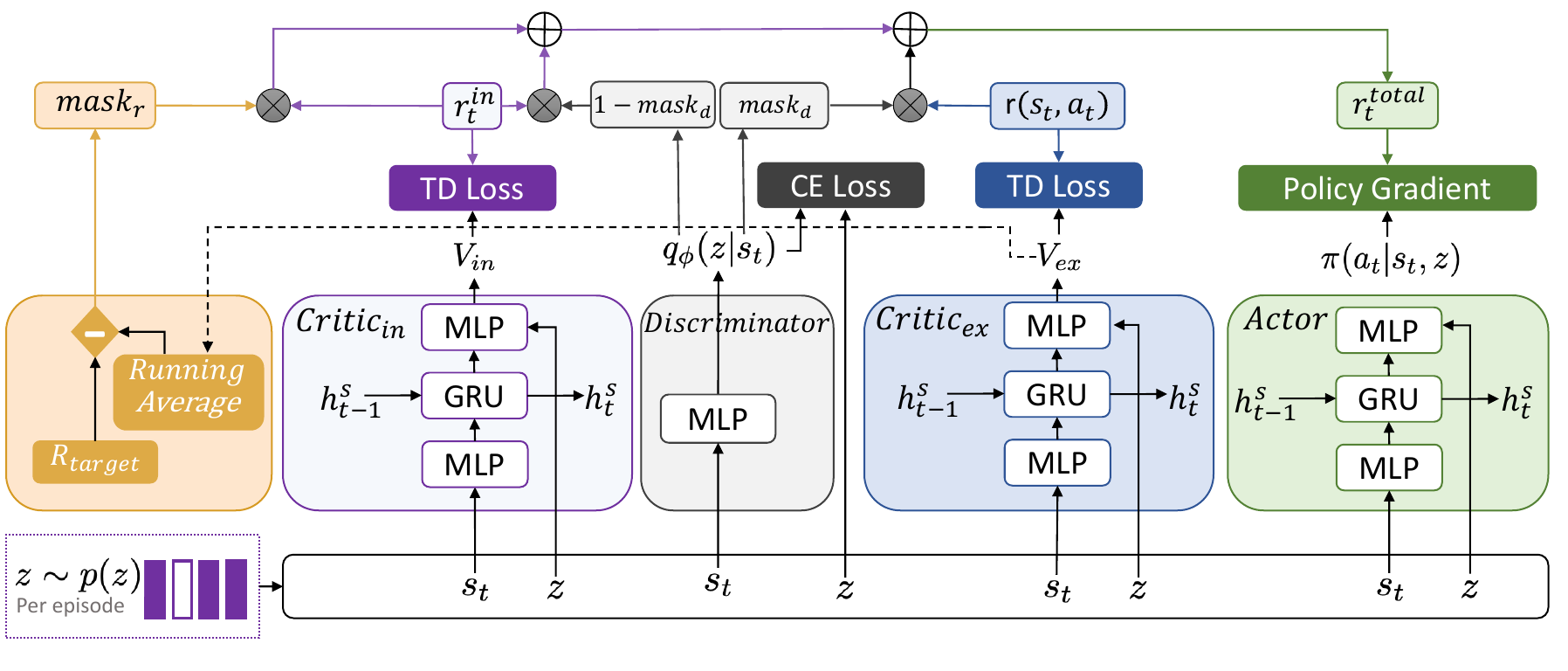}
  
  \caption{The overall framework of the DGPO algorithm. {\bf Top} illustrates the way of calculating $r^{total}_t$, where $mask_r=\mathds{I}[J(\theta) \geq R_{target}]$ and $mask_d=\mathds{I}[J_{Div}(\theta)\geq\delta]$. {\bf Center} shows the network structure and the data flow of the DGPO algorithm. {\bf Bottom} shows the latent variable sampling process.}
  \label{fig: network structure}
\end{figure*}

Strategies that solve a given RL task may be very distinct. One can think of these as a set of discrete points in $\rho^{\pi}(s)$ space -- if the distance between the points is large, perturbing around one single solution may not allow the discovery of all optimal strategies. Thus, we formulate the policy optimization process as a diversity-constrained optimization problem,
\begin{eqnarray}
\begin{aligned}
\max_{\pi_\theta} J(\theta),\ \mathrm{s.t.}\  J_{Div}(\theta) \geq \delta.
\end{aligned}
\label{eq: Diversity-Constrained optimization}
\end{eqnarray}

\begin{figure*}[t]
\begin{center}
\subfloat[]{\begin{centering}
\includegraphics[width=0.48\linewidth]{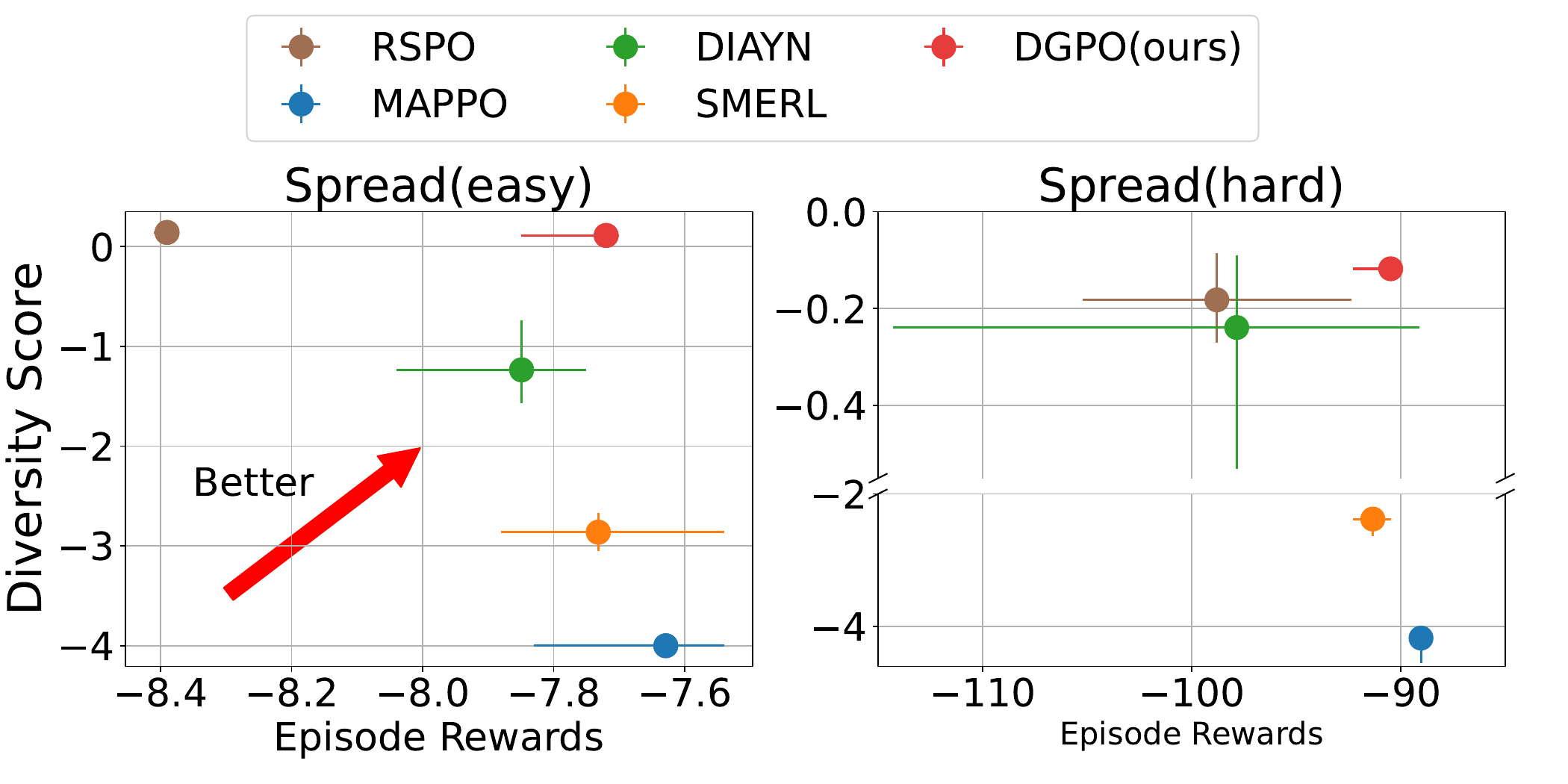}
\end{centering}
}
\subfloat[]{\begin{centering}
\includegraphics[width=0.48\linewidth]{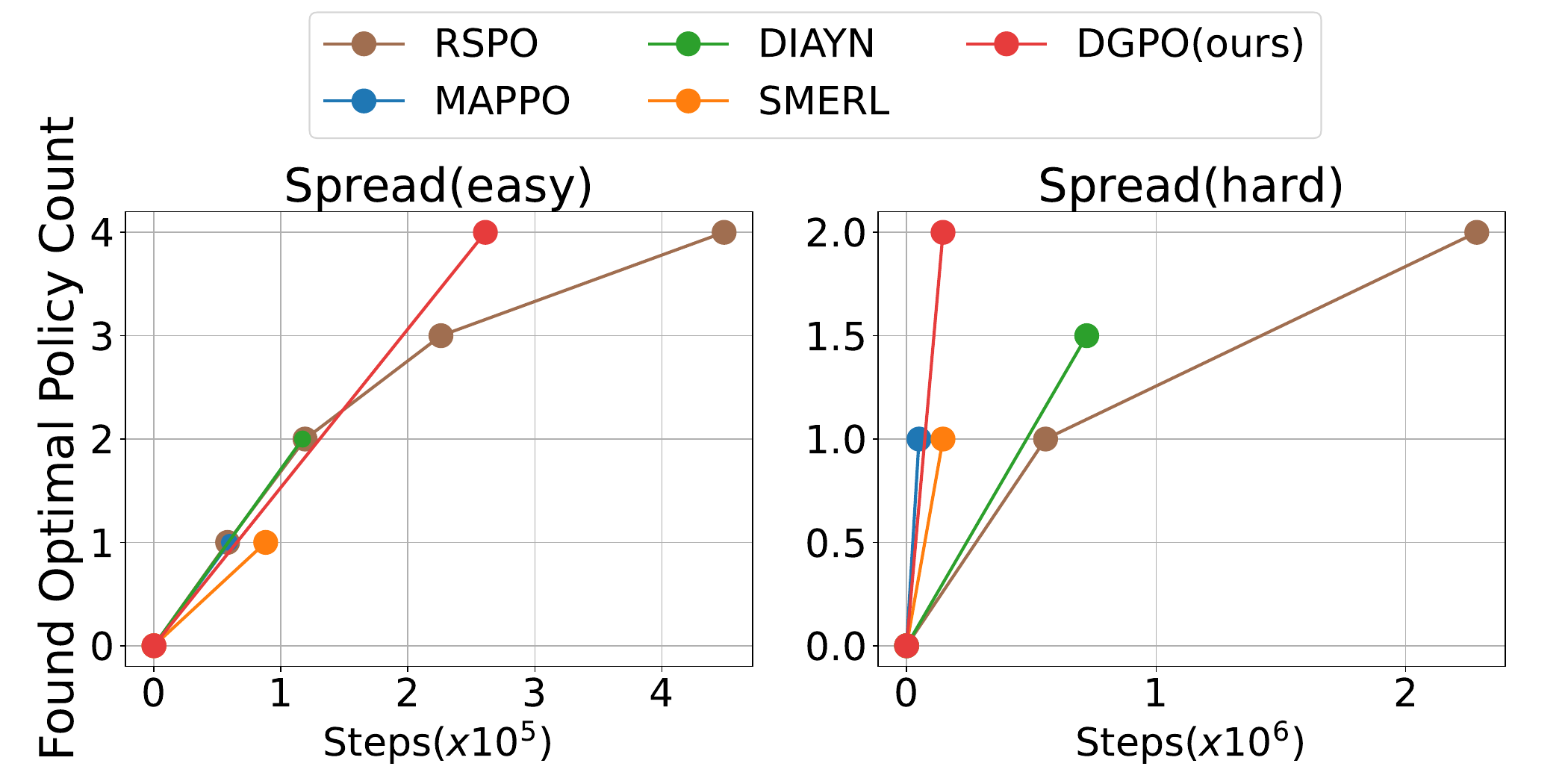}
\end{centering}
}
\end{center}
\caption{Experimental results in two MPE scenarios -- Spread (easy) and Spread (hard) -- each with multiple optimal solutions. 
(a) Plot showing extrinsic reward performance vs. how diverse the set of discovered strategies are. Positions in the upper–right corner are preferred -- DGPO is located here.
(b) Plot showing at which point in training each optimal strategy is discovered. Results show that only DGPO and RSPO can find all the solutions. But DGPO achieved {\bf over $1.7\times$} and {\bf $15\times$ speedup} in convergence speed compared to RSPO in the Spread (easy) and Spread (hard) scenarios, respectively.
}
\label{fig: MPE_result}

\end{figure*}

Under this objective, individual policies are constrained to keep a certain distance $\delta$ apart from each other, and systematically explore their own regions.
We can introduce a Lagrange multiplier $\lambda$ to tackle this constrained optimization problem,
\begin{eqnarray}
\begin{aligned}
&\max_{\pi_\theta}\min_{\lambda\geq0}  J(\theta) + \lambda (J_{Div}(\theta)-\delta) \\
\geq& \max_{\pi_\theta}\mathbb{E}_{z\sim p(z),s\sim\rho^{\pi}(s)} \\
&\left[\min_{\lambda\geq0} \mathbb{E}_{a\sim\pi(\cdot|s,z)}[\sum_t\gamma^t r(s_t,a_t)]+\lambda(\sum_t \gamma^t r^{in}_t-\delta)\right].
\end{aligned}
\label{eq: Lagrange multiplier}
\end{eqnarray}
The proof can be found in Appendix E. Eq.~\ref{eq: Lagrange multiplier} provides a lower bound on the Lagrange multiplier objective, which can be optimized more easily than the original problem. Eq.~\ref{eq: Lagrange multiplier} can be interpreted as optimizing the extrinsic-rewards, $J(\theta)$, when diversity is greater than some threshold. Otherwise, the intrinsic rewards objective $J_{Div}(\theta)$ are optimized. From another perspective, one can think of masking out terms in elements in Eq.~\ref{eq:probabilistic inference} based on a diversity metric. The updated objective can be written,
\begin{eqnarray}
\begin{aligned}
& p(\mathcal{O}_t=1|s_t,a_t,z) =
\exp\left(\mathbb{I}[J_{Div}(\theta)\geq\delta] r(s_t,a_t)\right), \\
& p(z|s_t,a_t) = 
\exp\left((1-\mathds{I}[J_{Div}(\theta)\geq\delta])r^{in}_t\right),
\end{aligned}
\label{eq:Diversity-term in First-Stage}
\end{eqnarray}
where $\mathds{I}[ \cdot ]$ is the indicator function.

\subsection{Stage 2: Extrinsic-Reward-Constrained Optimization}
The objective developed in the previous section can return a set of discrete optimal points. However, sometimes two strategies may still converge to the same sub-optimal point. This can destabilize the training process since both strategies are attracted to the same optimal point but simultaneously repelled by each other. To stabilize the training process and improve diversity, we relax the definition of ``optimal'' and assume that a policy with accumulated extrinsic rewards greater than some target value $R_{target}$ is an optimal policy. Thus, the policy optimization process can be formulated as an extrinsic-reward-constrained optimization problem,
\begin{eqnarray}
\begin{aligned}
\max_{\pi_\theta} J_{Div}(\theta), \ \text{s.t.}\  J(\theta)\geq R_{target}.
\end{aligned}
\label{eq:Extrinsic-Reward-Constrained optimization}
\end{eqnarray}

This means that if two strategies try to converge to the same sub-optimal point. They are allowed to find their destiny in the neighborhood of the optimal point to further maximize the level of diversity. 
On the other hand, for those policies that are already sufficiently distinct, the diversity objective serves as an intrinsic reward to encourage exploration. Similar to how we deal with diversity-constrained optimization. We implement Eq.~\ref{eq:Extrinsic-Reward-Constrained optimization} by injecting an extrinsic-rewards-constraint into the framework. The updated elements in Eq.~\ref{eq:probabilistic inference} can be defined as: 
\begin{eqnarray}
\begin{aligned}
&p(\mathcal{O}_t=1|s_t,a_t,z) = \exp(r(s_t,a_t)), \\
&p(z|s_t,a_t) = \exp(\mathds{I}[J(\theta) \geq R_{target}]r^{in}_t).
\end{aligned}
\label{eq:R_total}
\end{eqnarray}

\subsection{Diversity-Guided Policy Optimization}

In this section, we will introduce our final algorithm which unifies the two-stage training processes into one unified algorithm and also introduce how to implement it with the on-policy RL algorithm. We develop a new variation of PPO~\cite{schulman2017proximal} by considering policy network and critic network that are conditioned on latent variable $z$, i.e., $\pi(a_t|s_t,z)$. The critic network is divided into two parts, i.e., $V^{\pi}_{\psi_{ex}}(o_{1:t},z)$  and $V^{\pi}_{\psi_{in}}(o_{1:t},z)$, where $\psi_{ex}$ and $\psi_{in}$ are their parameters. The parameters of critic networks can be trained by a temporal difference (TD) loss~\cite{sutton2018reinforcement}:
\begin{eqnarray}
\begin{aligned}
&L(\psi_{ex}) = \\
&MSE\left(V^{\pi}_{\psi_{ex}}(o_{1:t},z), \sum_{t'=t}^\infty \gamma^{t'-t} \mathds{I}[J_{Div}(\theta)\geq\delta]r(s_{t'},a_{t'})\right), \\
&L(\psi_{in}) = MSE \Bigg(V^{\pi}_{\psi_{in}}(o_{1:t},z), \\
&\sum_{t'=t}^\infty \gamma^{t'-t} \left[(1-\mathds{I}[J_{Div}(\theta)\geq\delta])+ \mathds{I}\left[J(\theta) \geq R_{target}\right]\right]r^{in}_{t'} \Bigg),
\end{aligned}
\label{eq:value_loss}
\end{eqnarray}

where $L(\cdot)$ stands for loss function and $MSE(\cdot)$ stands for mean square error. We maintain a running average of $V^{\pi}_{\psi_{ex}}(s_t,z)$ to approximate $\mathbb{E}_{s\sim\rho^{\pi}(s), a\sim\pi(\cdot|s)}[\sum_t\gamma^t r(s_t,a_t)]$. We also construct a discriminator $q_\phi(z|s_t)$ that takes the state as input and predict the probability of latent variable $z$, where $\phi$ is the parameter of the discriminator network. And the discriminator can be trained in a supervised manner:
\begin{eqnarray}
\begin{aligned}
L(\phi) = \mathbb{E}_{(s_t,a_t,z)\sim \mathcal{D}_\pi}[CE(q_\phi(s_t),z)],
\end{aligned}
\label{eq:discriminator_loss}
\end{eqnarray}
where $CE(\cdot,\cdot)$ stands for cross entropy loss. We implement DGPO by incorporating diversity-constrained optimization and extrinsic-reward-constrained optimization into the same framework, i.e., the total value of the state and the total reward can be defined as below:
\begin{eqnarray}
\begin{aligned}
&V^{\pi}_{total}(o_{1:t},z) = 
V^{\pi}_{\psi_{in}}(o_{1:t},z) +V^{\pi}_{\psi_{ex}}(o_{1:t},z), \\
&r_t^{total} = \mathds{I}[J_{Div}(\theta)\geq\delta]r(s_{t'},a_{t'}) \\
& \qquad + \left[(1-\mathds{I}[J_{Div}(\theta)\geq\delta])+\mathds{I}\left[J(\theta) \geq R_{target}\right]\right]r^{in}_{t'}.
\end{aligned}
\label{eq:r_total}
\end{eqnarray}

 In theory, it is not feasible to simultaneously conduct diversity-constrained optimization and extrinsic-reward-constrained optimization. As a result, the aforementioned implementation serves as an approximation to the original objective. We posit that this approximation is reasonable, as the empirical findings demonstrate that the two training stages are not concurrent. Fig.~\ref{fig: network structure} shows the overall framework of the DGPO algorithm. The detailed training process of DGPO can be found in the Appendix G.


\begin{figure*}[t]
\begin{center}
\subfloat[]{\begin{centering}
\includegraphics[width=0.48\linewidth]{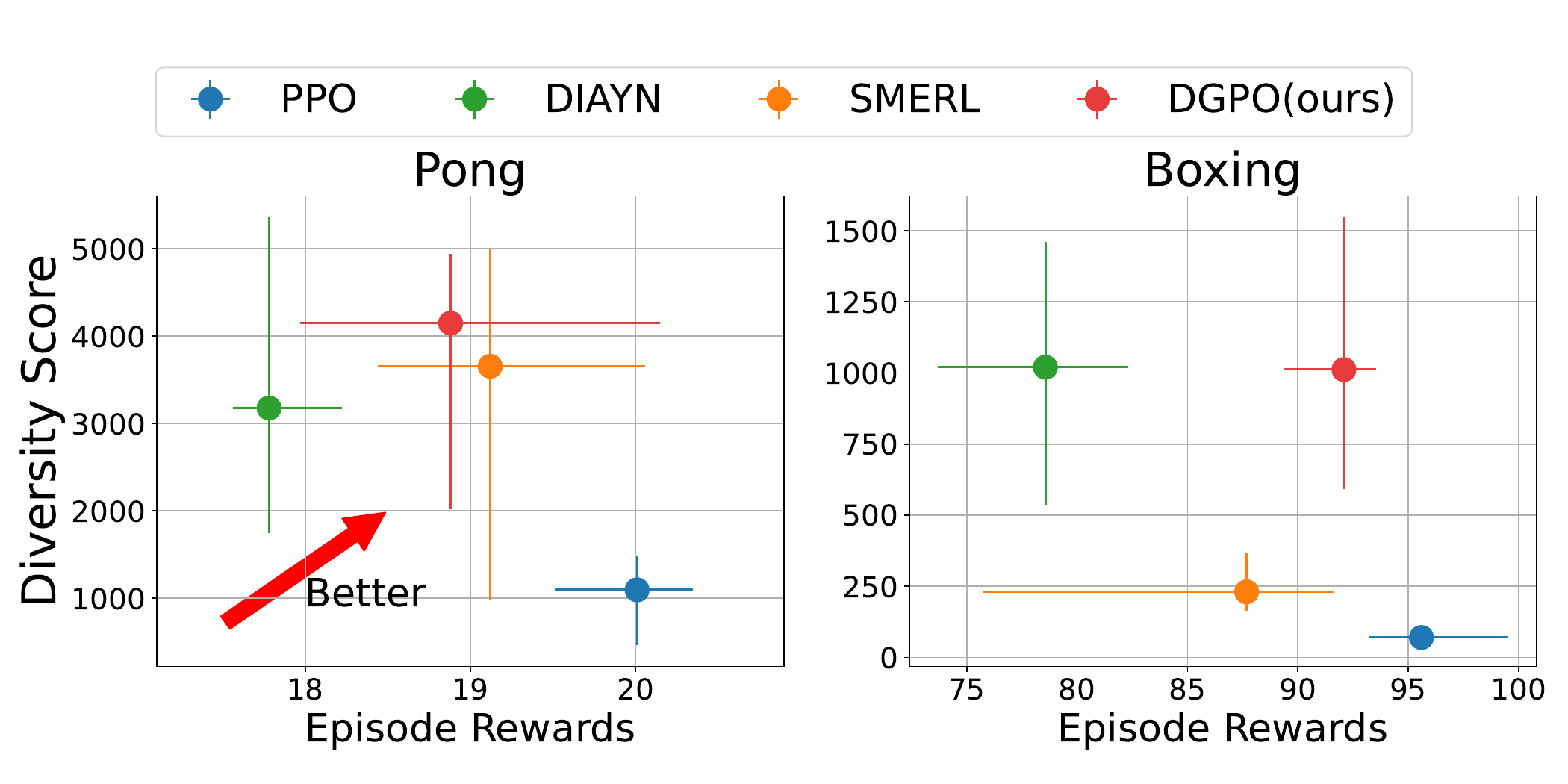}
\end{centering}
}
\subfloat[]{\begin{centering}
\includegraphics[width=0.48\linewidth]{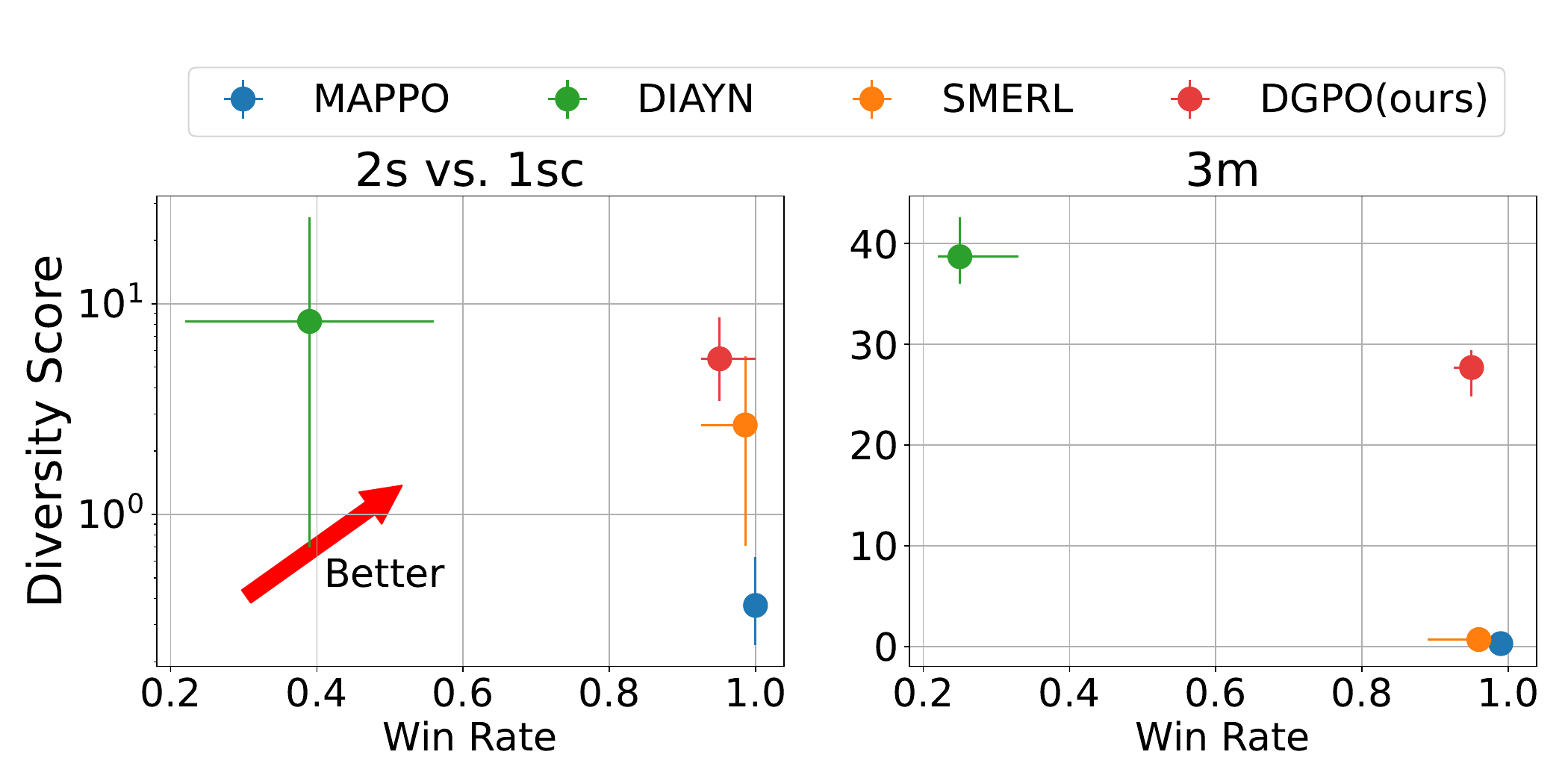}
\end{centering}
}
\end{center}

\caption{Plots showing extrinsic reward performance vs. the diversity of the set of discovered strategies. (a) In two Atari games. (b) In two SMAC scenarios.}
\label{fig:Atari_SMAC_result}

\end{figure*}

\section{Experiments}

In this section, we evaluate our algorithm on several RL benchmarks -- Multi-agent Particle Environment (MPE)~\cite{mordatch2018emergence}, StarCraft Multi-Agent Challenge (SMAC)~\cite{samvelyan2019starcraft}, and Atari~\cite{bellemare2013arcade}. We compare our algorithm to four baseline algorithms:
\\
{\bf MAPPO}~\cite{yu2021surprising}: MAPPO adapts the single-agent PPO~\cite{schulman2017proximal} algorithm to the multi-agent setting by using a centralized value function with shared team-based rewards.
\\
{\bf DIAYN}~\cite{eysenbach2018diversity}: DIAYN trains agents with a mutual-information based intrinsic reward to discover a diverse set of skills. In our setup, these intrinsic rewards are combined with extrinsic rewards.
\\
{\bf SMERL}~\cite{kumar2020one}: SMERL maximizes a weighted combination of intrinsic rewards and extrinsic rewards when the return of extrinsic reward is greater than some given threshold.
\\
{\bf RSPO}~\cite{zhou2022continuously}: RSPO is an iterative algorithm for discovering a diverse set of quality strategies. It toggles between extrinsic and intrinsic rewards based on a trajectory-based novelty measurement.

As far as possible, all methods use the same hyper-parameters. However, there is some difference for RSPO, for which we use the open-source implementation. (Full hyperparameters are listed in the Appendix B.) All experiments were performed on a machine with 128 GB RAM, one 32-core CPU, and one GeForce RTX 3090 GPU.


\subsection{Multi-Agent Particle Environment}

\begin{figure}[h]
\begin{center}
\subfloat[Easy]{\begin{centering}
\includegraphics[width=0.25\linewidth]{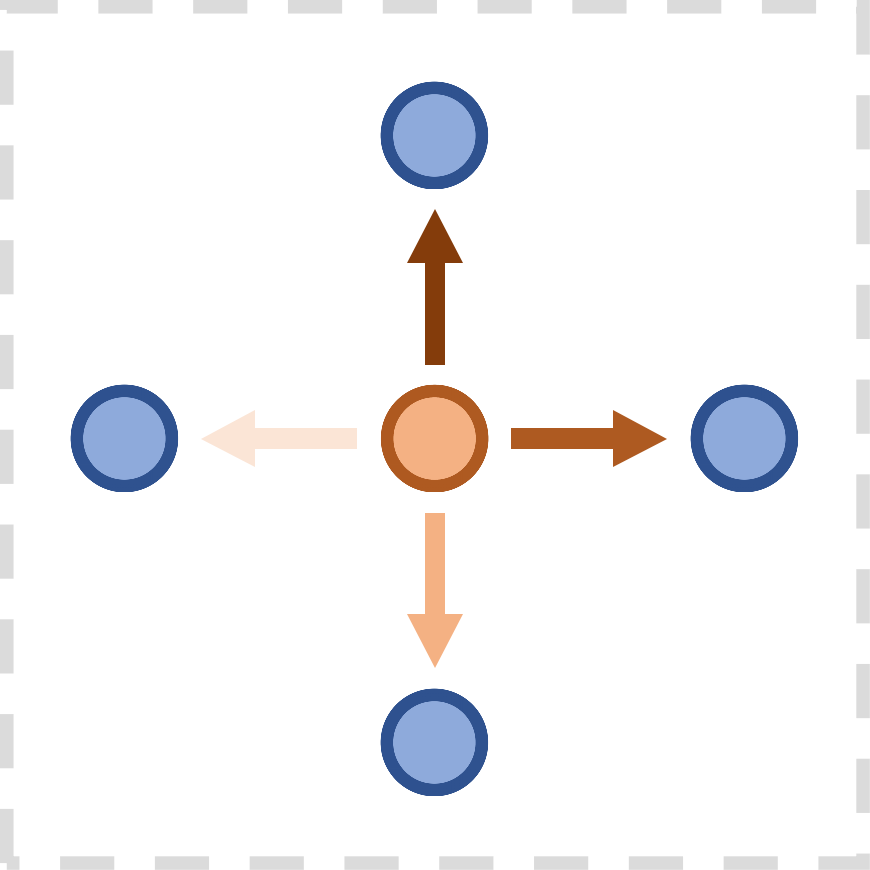}
\end{centering}
}
\subfloat[Hard]{\begin{centering}
\includegraphics[width=0.25\linewidth]{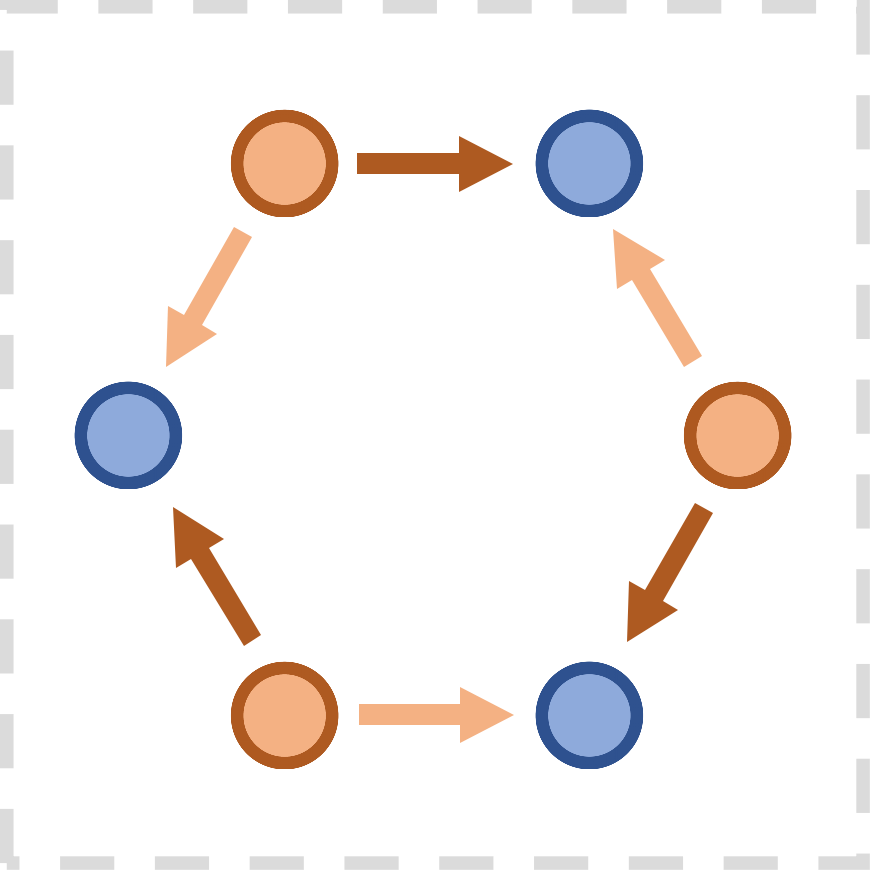}
\end{centering}
}
\end{center}
\caption{The initial state of Spread (easy) and Spread (hard). In both scenarios, Agents (orange dots) aim to reach one of the destinies (blue dots). We highlight the optimal solutions with arrows of different colors.}
\label{fig: MPE_maps} 
\end{figure}

We evaluate on two scenarios shown per Fig.~\ref{fig: MPE_maps}, Spread (easy) and Spread (hard). In Spread (easy), there are four landmarks and one agent. The agent starts from the center and aims to reach one of the landmarks, giving four optimal solutions. In Spread (hard), there are three agents and three landmarks. 
Agents cooperate to cover all the landmarks and avoid colliding with others, giving two optimal solutions. 
Model weights are shared across agents.

\begin{figure}[h]
\begin{center}
\subfloat[Pong]{\begin{centering}
\includegraphics[width=0.9\linewidth]{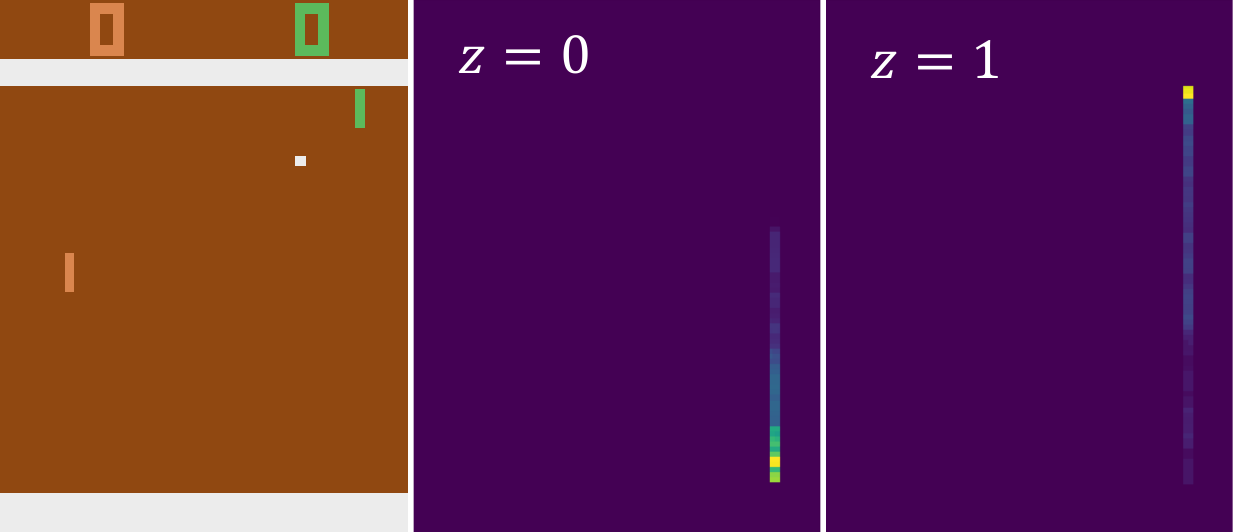}
\end{centering}
}
\vspace{0.1mm}
\subfloat[Boxing]{\begin{centering}
\includegraphics[width=0.9\linewidth]{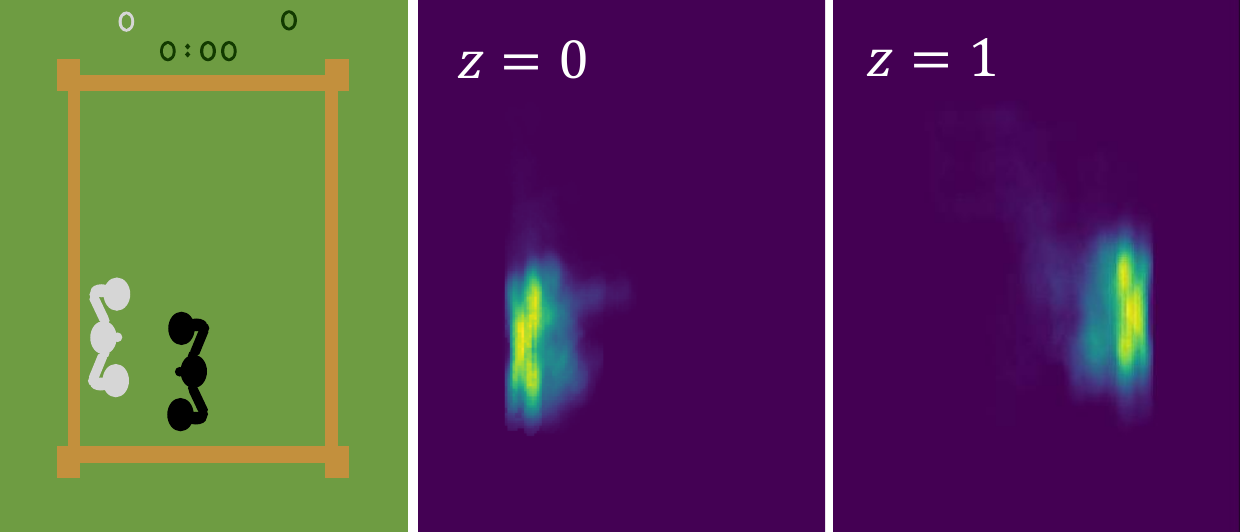}
\end{centering}
}
\end{center}
\caption{Screen shots and heat maps of agents' trajectories on Pong and Boxing. (a)  In the Pong game, our agent controls the paddle (green block) to hit the ball (white dot). 
(b) In the Boxing game, our agent (in white) has a boxing match with the opponent (in black).
}
\label{fig: atari visualization}

\end{figure}

Similar to \cite{parker2020effective}, we introduce a metric to quantitatively evaluate the diversity score of the given set of policies, $\Pi$: $M_{Div}(\Pi) = \frac{1}{\mathit{n_z}}\sum_{i=1}^{\mathit{n_z}}\sum_{j=i+1}^{\mathit{n_z}}\ln(\Vert\Phi(\pi_i)-\Phi(\pi_j)\Vert_2)$,
where $\Phi(\pi)$ is the behavior embedding of the policy $\pi$. For MPE, $\Phi(\pi)$ is represented by the concatenated agents' positions over an episode. 

We set $n_z=4$ in Spread (easy) and $n_z=2$ in Spread (hard) to test whether an algorithm can discover all optimal solutions. 
The performance, diversity score, and steps required for convergence for all algorithms are given in Fig.~\ref{fig: MPE_result}. Results are averaged over five seeds.

DGPO provides a favorable combination of high diversity, and high reward. It also exhibits rapid convergence.
MAPPO only recovers a single solution in each setting. 
DIAYN only finds 2/4 optimal solutions in Spread (easy) and often only 1/2 in Spread (hard). This supports our earlier claim that a naive combination of extrinsic and intrinsic rewards is insufficient.
SMERL is only able to discover new strategies by perturbing around a discovered global optimal. Thus, it is unable to find more than one optimal strategy. 
RSPO is the only other method also to recover all optimal strategies. However, relative to DGPO, RSPO achieves lower overall reward and slower convergence. 



\subsection{Atari}

We evaluate DGPO on the Atari games Pong and Boxing, to test the performance for tasks with image observations. DGPO's diversity metric is defined similarly to the one used in MPE (full details in Appendix F). In each environment, we set $n_z=2$.
Results are summarized in Fig.~\ref{fig:Atari_SMAC_result}(a), averaged over five seeds. We also reported the experimental results for $n_z=10$ in Appendix I. DGPO again delivers a favorable trade-off between external reward and strategy diversity. Fig.~\ref{fig: atari visualization}(a) visualizes the results of the two strategies obtained by DGPO on Pong. In this game, the agent controls the green paddle. When $z=0$ the agent holds the paddle at the bottom of the screen until the ball is near, while when $z=1$, the default position of the paddle is at the top of the screen. 
Fig.~\ref{fig: atari visualization}(b) similarly shows the different strategies obtained by DGPO on Boxing, where the agent controls the white character, and is rewarded for punching the black opponent. The heatmap shows DGPO learns to attack from different sides. Other baseline algorithms tend to remain in a single corner. 

\begin{figure}
  \centering
  \includegraphics[width=0.91\linewidth]{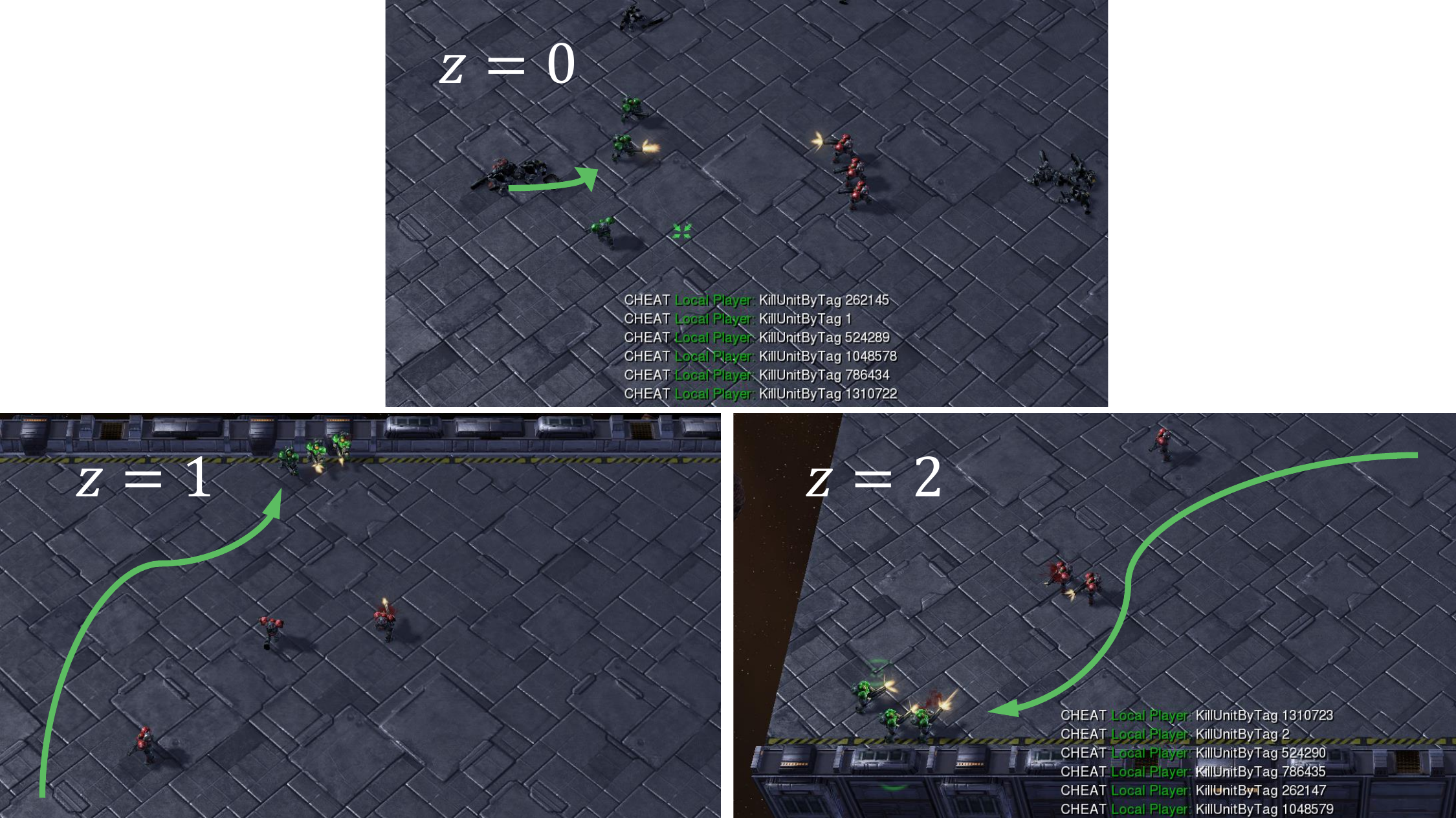}
  \caption{Visualization results of the three strategies obtained by DGPO on $\textit{3m}$ map. The green arrows show the trajectories of our agents (in green). 
  }
  \label{fig:smac_exp}
\end{figure}

\subsection{StarCraft II}

We conduct experiments on two StarCraft II maps ($\textit{2s\_vs.\_1sc}$ \& $\textit{3m}$) from SMAC. We set $n_z=3$ and measure the mean win rates over five seeds. Fig.~\ref{fig:Atari_SMAC_result}(b) shows that, relative to other algorithms, DGPO discovers sets of strategies that are both diverse and achieve good win rates. 
Fig.~\ref{fig:smac_exp} visualizes three strategies obtained by DGPO on $\textit{3m}$ map. In this map, we control the three green agents to combat the red built-in agents. We visualize the trajectories of our agents with green arrows. When $z=0$, the policy produces an aggressive strategy, with agents moving directly forward to attack the enemies. When $z=1$, the agents display a kiting strategy, alternating between attacking and moving. This allows them to attack enemies while limiting the taken damage. When $z=2$, the policy produces another kiting strategy but now with a downwards, rather than upwards drift.

\begin{figure}[t]
\begin{center}
\subfloat[]{\begin{centering}
\includegraphics[width=0.725\linewidth]{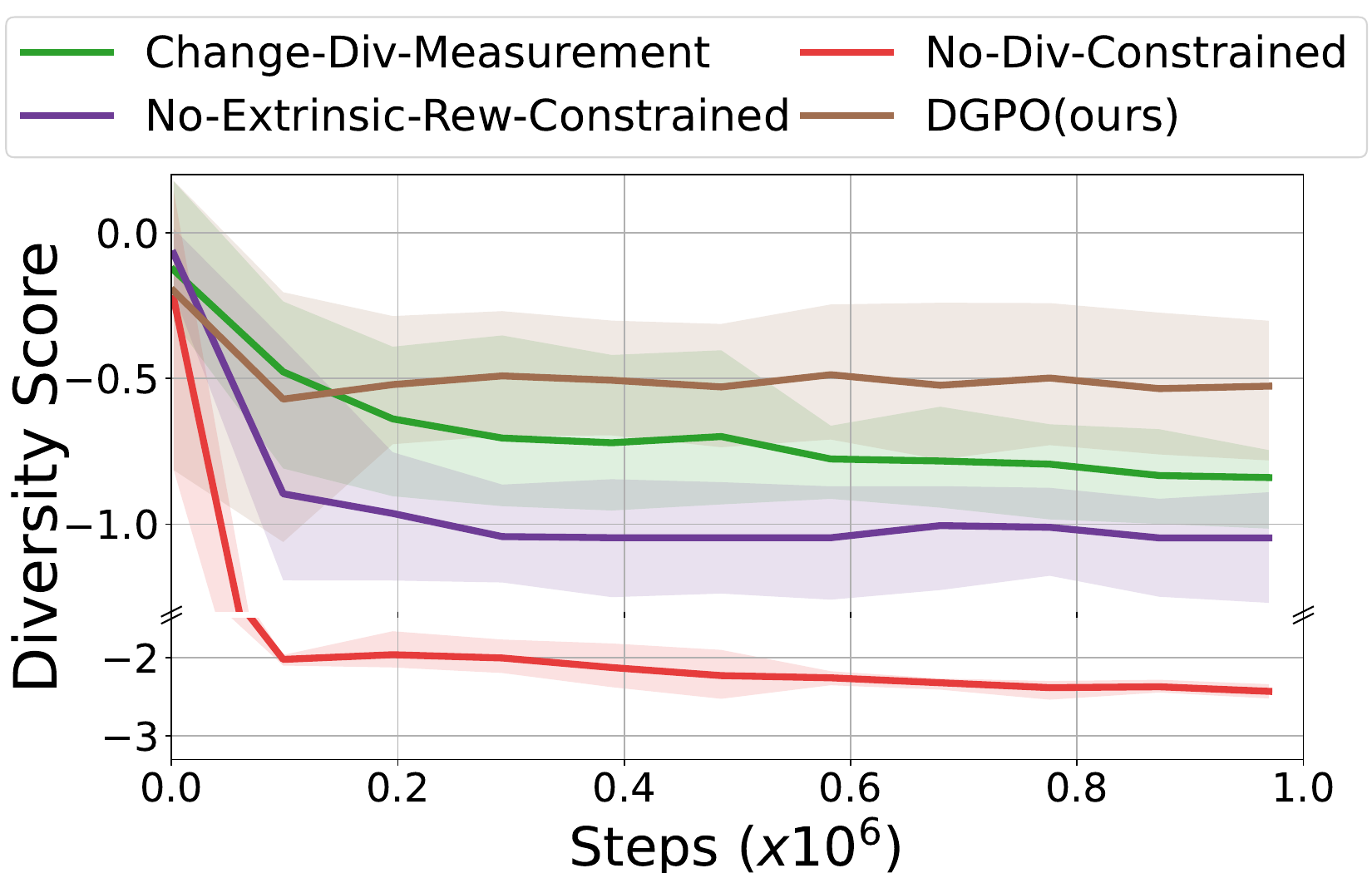}
\end{centering}
}
\vspace{0.1mm}
\subfloat[]{\begin{centering}
\includegraphics[width=0.725\linewidth]{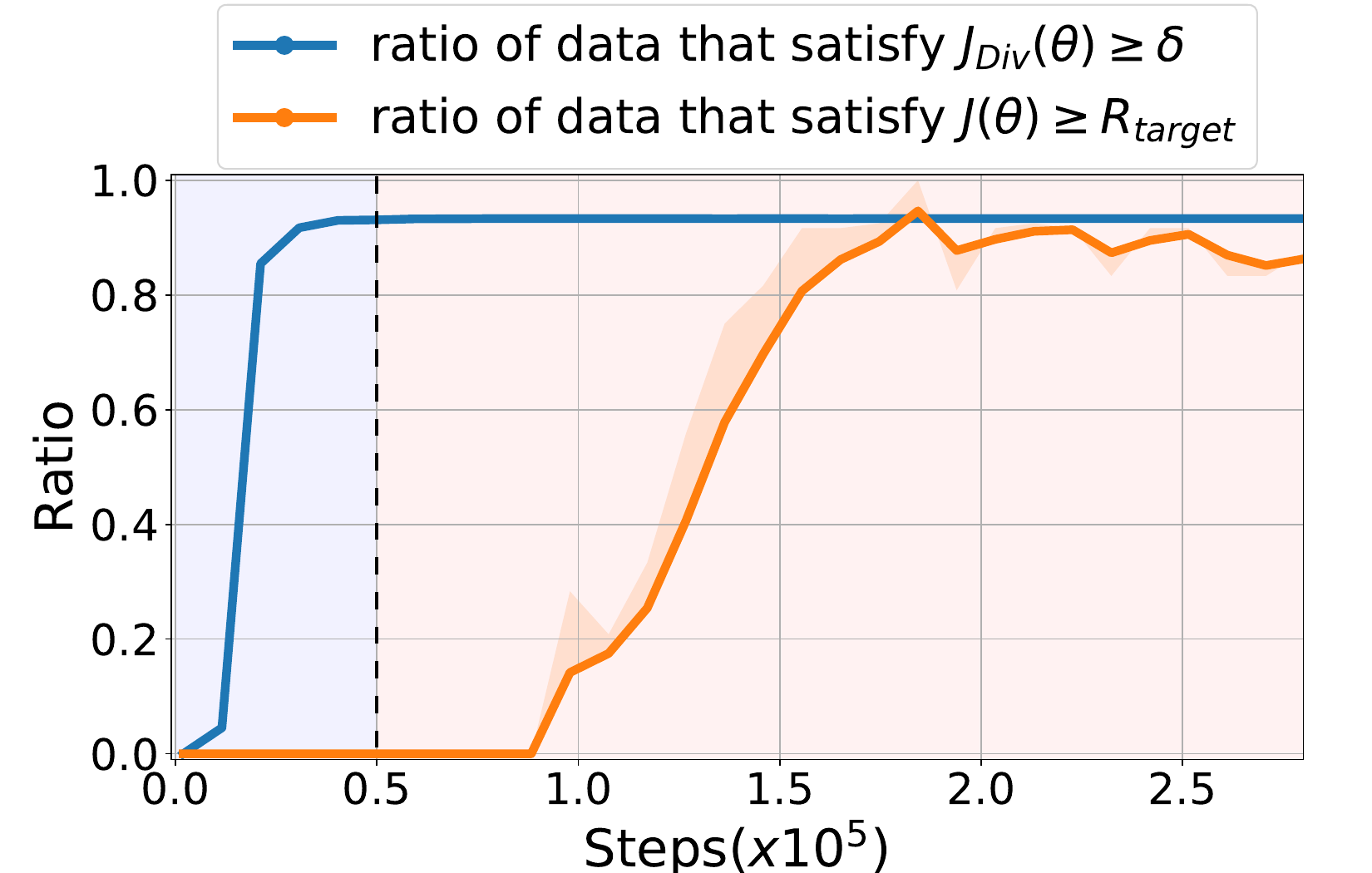}
\end{centering}
}
\end{center}
\caption{(a) The impact of each component of our algorithm on diversity scores in the MPE Spread (hard) scenario. (b) DGPO can be distinctly divided into two stages: diversity-constrained optimization and extrinsic-reward-constrained optimization.}
\label{fig:Ablation}

\end{figure}

\subsection{Ablation Study}

We performed ablation studies on MPE Spread (hard) tasks, systematically removing each element of our algorithm to assess its impact on diversity. The empirical result is shown in Fig.~\ref{fig:Ablation}(a). Throughout this section, we set $n_z=3$ and the result is averaged over 5 seeds.
{\bf Change-Diversity-Measurement} uses mutual information as shown in Eq.~\ref{eq:Diversity-metric-1} as diversity metric. 
Experimental results indicate that in the Spread (hard) scenario, the diversity score of Change-Diversity-Measurement is lower than that of DGPO. While optimizing three policies simultaneously, the limited number of optimal solutions (only two) leads to one policy behaving differently while the other two exhibit similar behavior. Consequently, the diversity score based on mutual information becomes artificially high (as it reflects the overall diversity level of the policy set), causing the policy to stop optimizing diversity, even though two policies continue to behave similarly.
{\bf No-Diversity-Constrained-Optimization} excludes diversity-constrained optimization. Empirical results reveal that it only identifies one optimal solution. This suggests that utilizing the diversity metric as a constraint, rather than blending it with extrinsic rewards during the initial stages of training, significantly enhances the algorithm's effectiveness.
{\bf No-Extrinsic-Reward-Constrained-Optimization} omits extrinsic-reward-constrained optimization. 
While DGPO continues to enhance the diversity score as the expected return surpasses $R_{target}$, No-Extrinsic-Reward-Constrained-Optimization fails to do so. Consequently, it ultimately attains a lower diversity score compared to DGPO.

Our algorithm can be divided into two distinct stages, as depicted in Fig.~\ref{fig:Ablation}(b).
In the initial stage, we focus on diversity-constrained optimization until policies exhibit sufficient behavioral differences. Once the expected return reaches $R_{target}$, we transition to the extrinsic-reward-constrained optimization stage. Here, we maintain a fixed performance level at $R_{target}$ while maximizing the diversity score. These stages are clearly separated and non-overlapping.


\section{Conclusions}
\label{sec:conlusion}
In conclusion, this paper introduced the Diversity-Guided Policy Optimization (DGPO) algorithm, which demonstrates its capability to efficiently uncover diverse strategies that yield high rewards. By framing the training process as a pair of constrained optimization problems and solving them through probabilistic inference, DGPO stands out as a promising on-policy algorithm. Through experiments, we observed that DGPO strikes a favorable balance between diversity scores and rewards, all while exhibiting improved sample efficiency. Moving forward, we envision delving deeper into its potential to handle challenges such as exploration, self-play, and robustness.

\medskip

\bibliography{aaai24}

\clearpage

\appendix

\section{A: Network structure for Atari Games}
\label{ap:Atari_Network_structure}
\begin{figure}[h]
  \centering
  \includegraphics[width=0.98\linewidth]{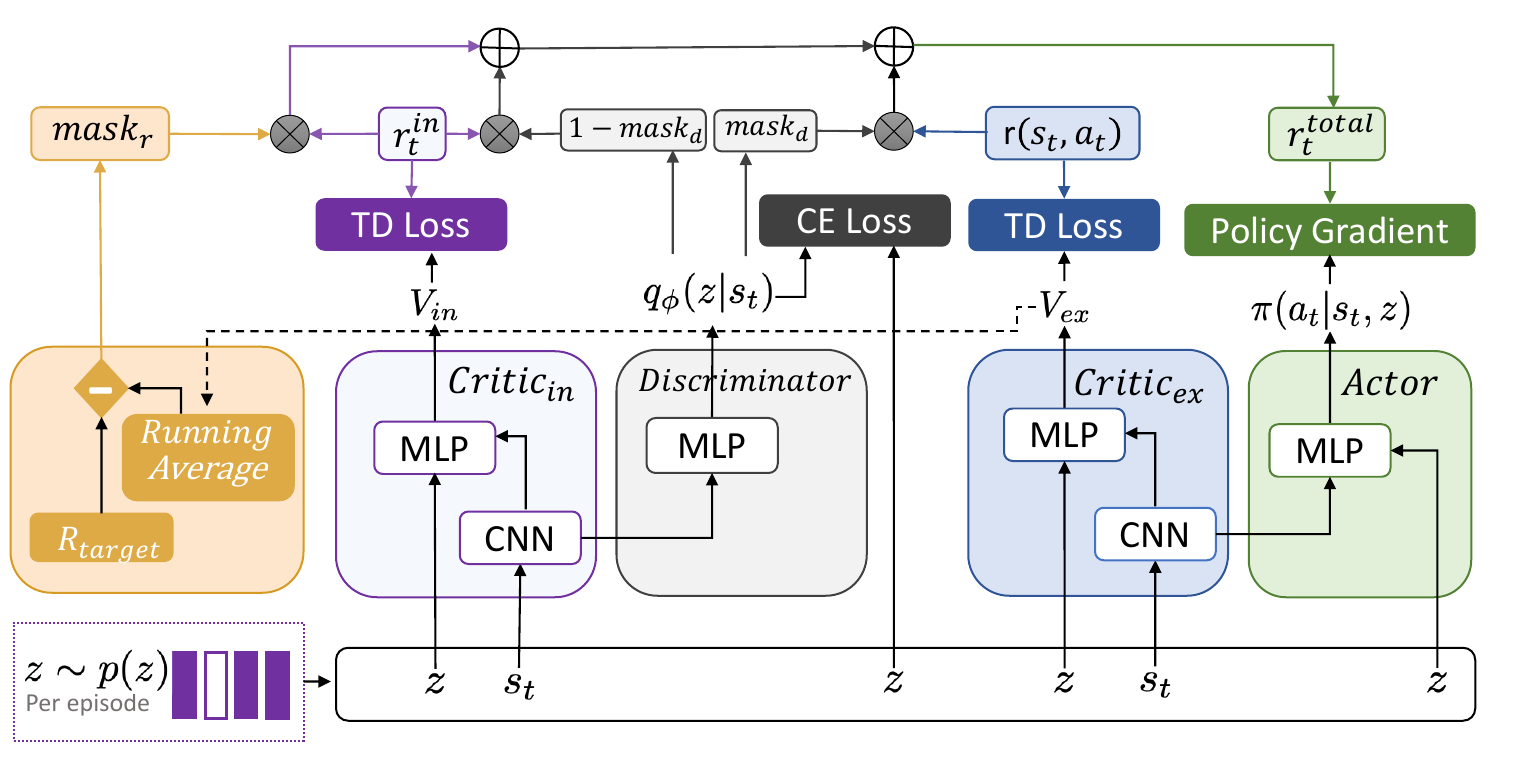}
  \vspace{-2mm}
  \caption{The overall framework of the DGPO algorithm for Atari games.}
  \label{fig: network structure for Atari}
  \vspace{-6mm}
\end{figure}

The overall framework of the DGPO algorithm for Atari games is summarized in Figure~\ref{fig: network structure for Atari}. We introduce two Convolutional Neural Networks(CNN) to deal with image inputs. One for extracting features for diversity and another for extracting features for calculating extrinsic rewards.

\section{B: Experimental Hyper-Parameters}
\label{ap:hyper_parameters}

\begin{table}[h]
\begin{center}
    \begin{tabular}{cccc}
       \toprule
       Common Hyper-Parameters     & MPE & SMAC & Atari \\
      \hline
num rollout threads & 128 & 30 & 16 \\
episode length & 15 & 400 & 128 \\
ppo epoch & 10 & 10 & 4 \\
   \bottomrule
\end{tabular}
\end{center}
  \caption{Hyper-parameters used in MAPPO, DIAYN, SMERL and DGPO across all domains.}
  \label{table:common_MPE}
  \vspace{-0.5cm}
\end{table}

\begin{table}[h]
\begin{center}
    \begin{tabular}{cc}
       \toprule
      Common Hyper-Parameters     & Value \\
      \hline
      recurrent data chunk length & 10\\
      max clipped value loss      & 0.2\\
      gradient clip norm          & 10.0\\
      gae lamda                   & 0.95\\
      gamma                       & 0.99\\
      value loss                  & huber loss\\
      huber delta                 & 10.0\\
      num GRU layers & 1 \\
      RNN hidden state dim & 64 \\
      fc layer dim & 64 \\
      fc layer num & 1 \\
      lr & 5e-4 \\
      discriminator lr & 1e-4\\
      gain & 0.01 \\
       \bottomrule
    \end{tabular}
  \end{center}
  \caption{Common hyper-parameters used in MAPPO, DIAYN, SMERL and DGPO across all domains.}
  \label{table:common_hyper}
  \vspace{-0.5cm}
\end{table}

\begin{table}[h]
\begin{center}
\begin{tabular}{ccc}
\toprule
scenario & likelihood-alpha & likelihood-threshold \\
\hline
Spread(easy) & & \\
iteration1 & 0.5 & 70 \\
Spread(easy) & & \\ 
iteration2$\sim$4 & 0.995 & 70 \\
Spread(hard) & & \\
iteration1 & 0.5 & 40 \\
Spread(hard) & & \\
iteration2$\sim$4 & 0.995 & 40 \\
\bottomrule
\end{tabular}
\end{center}
  \caption{Hyper-parameters used in RSPO across all domains.}
  \label{table:RSPO_para}
  \vspace{-0.5cm}
\end{table}

\begin{table}[h]
\begin{center}
\begin{tabular}{ccccccc}
\toprule
scenario & Spread(easy) & Spread(hard) & Pong \\
\hline
diversity thresh & log(0.9) & log(0.9) & log(0.8) \\
$R_{ex}$ thresh & -2.5 & -37 & 0.8 \\ 
$n_z$ & 4 & 2 & 2 \\ 
\bottomrule
scenario & Boxing & 2s vs. 1sc & 3m \\
\hline
diversity thresh & log(0.8) & log(0.8) & log(0.8) \\
$R_{ex}$ thresh & 6.5 & 11.5 & 13.5 \\ 
$n_z$ & 2 & 3 & 3 \\ 
\bottomrule
\end{tabular}
\end{center}
  \caption{Hyper-parameters used in DGPO across all domains.}
  \label{table:common_DGPO}
  \vspace{-0.5cm}
\end{table}


\section{C: Derivation Process of the ELBO}
\label{ap:ELBO_proof}
\begin{eqnarray}
\begin{aligned}
&{\rm log}p(\mathcal{O}_{1:T}) \\
&=\log \int\int\int p(\mathcal{O}_{1:T},s_{1:T},a_{1:T},z) \mathit{d}s_{1:T}\mathit{d}a_{1:T}\mathit{d}z \\
&=\log \int\int\int p(\mathcal{O}_{1:T},s_{1:T},a_{1:T},z) \\
&\cdot\frac{q(s_{1:T},a_{1:T},z)}{q(s_{1:T},a_{1:T},z)} \mathit{d}s_{1:T}\mathit{d}a_{1:T}\mathit{d}z \\
&=\log \mathbb{E}_{(s_{1:T},a_{1:T},z)\sim q(s_{1:T},a_{1:T},z)}\left[\frac{p(\mathcal{O}_{1:T},s_{1:T},a_{1:T},z)}{q(s_{1:T},a_{1:T},z)} \right] \\
&\geq \mathbb{E}_{(s_{1:T},a_{1:T},z)\sim q(s_{1:T},a_{1:T},z)}[\log{p(\mathcal{O}_{1:T},s_{1:T},a_{1:T},z)} \\
&-\log{q(s_{1:T},a_{1:T},z)} ].
\label{eq:base}
\end{aligned}
\end{eqnarray}

The inequality on the last line is obtained via Jensen inequality. The first term in Equation~\ref{eq:base} can be written as:
\begin{eqnarray}
\begin{aligned}
&p(\mathcal{O}_{1:T},s_{1:T},a_{1:T},z) \\
&=p(s_1)\prod_{t=1}^Tp(s_{t+1}|s_t,a_t)p(\mathcal{O}_t|s_t,a_t,z)p(z|s_t,a_t).
\label{eq:first_term}
\end{aligned}
\end{eqnarray}

The second term in Equation~\ref{eq:base} can be written as:
\begin{eqnarray}
\begin{aligned}
&q(s_{1:T},a_{1:T},z) \\
&=q(s_1)\prod_{t=1}^Tq(s_{t+1}|s_t,a_t)q(a_t|s_t,z).
\label{eq:second_term}
\end{aligned}
\end{eqnarray}

Refer to ~\cite{levine2018reinforcement}, we fix $q(s_1)=p(s_1)$ and $q(s_{t+1}|s_t,a_t)=p(s_{t+1}|s_t,a_t)$. Finally, Equation~\ref{eq:base} can be written as below:
\begin{eqnarray}
\begin{aligned}
&{\rm log}p(\mathcal{O}_{1:T}) \\
&\geq \mathbb{E}_{(s_{1:T},a_{1:T},z)\sim q(s_{1:T},a_{1:T},z)}[\log{p(\mathcal{O}_{1:T},s_{1:T},a_{1:T},z)} \\
& -\log{q(s_{1:T},a_{1:T},z)} ] \\
&= \mathbb{E}_{(s_{1:T},a_{1:T},z)\sim q(s_{1:T},a_{1:T},z)}[\log{p(\mathcal{O}_t|s_t,a_t,z)} \\
&+\log{p(z|s_t,a_t)}-\log{q(a_t|s_t,z)} ] \\
\label{eq:ELBO_final}
\end{aligned}
\end{eqnarray}

\section{D: Lower bound of the diversity objective}
\label{ap:Div_lower_bound}

In this section, we prove Eq. 4 in the main paper.
\begin{eqnarray}
\begin{aligned}
&\mathrm{DIV}(\pi_{\theta}) \\
&= \mathbb{E}_{z\sim p(z)}\left[\min_{z'\neq z} D_{KL}(\rho^{\pi_\theta}(s|z)||\rho^{\pi_\theta}(s|z'))\right] \\
&=\mathbb{E}_{z\sim p(z)}\left[\min_{z'\neq z}\mathbb{E}_{s\sim\rho^\pi(s|z)}\left[\log{\frac{p(s|z)}{p(s|z')}}\right]\right] \\
&\geq \mathbb{E}_{z\sim p(z)}\left[\min_{z'\neq z}\mathbb{E}_{s\sim\rho^\pi(s|z)}\left[\log{\frac{p(s|z)}{p(s|z')+p(s|z)}}\right]\right] \\
&= \mathbb{E}_{z\sim p(z)}\left[\min_{z'\neq z}\mathbb{E}_{s\sim\rho^\pi(s|z)}\left[\log{\frac{p(s,z)}{p(s,z')+p(s,z)}}\right]\right] \\
&\geq \mathbb{E}_{s\sim\rho^\pi(s|z),z\sim p(z)}\left[\min_{z'\neq z}\log{\frac{p(s,z)}{p(s,z')+p(s,z)}}\right] \\
&= \mathbb{E}_{s\sim\rho^\pi(s|z),z\sim p(z)}\left[\min_{z'\neq z}\log{\frac{p(z|s)}{p(z'|s)+p(z|s)}}\right].
\label{eq:diversity_lower_bound}
\end{aligned}
\end{eqnarray}

The reason the inequality is valid in the third row is due to the fact that $a / (b + c)$ is a monotone decreasing function of $c$. The justification for the fifth row stems from the convex nature of the function $\min(\cdot)$, combined with the utilization of Jensen's inequality. To solve the numerical problem, we add $p(s|z)$ to the denominator in the third row. To implement Equation~\ref{eq:diversity_lower_bound}, we use a learned discriminator $q_\phi(z|s)$ to approximate $p(z|s)$.

\section{E: Diversity-Constrained Optimization}
\label{ap:Diversity-Constrained Optimization}
In this section, we show optimization problems in Eq.8 and Eq.9 in the main paper are equivalent. The original problem can be written as:
\begin{eqnarray}
\max_{\pi_\theta} J(\theta),\ \mathrm{s.t.}\  J_{DIV}(\pi_\theta) \geq \delta.
\label{eq:constrained_optimization}
\end{eqnarray}

To further tight up the constraint, we replace $\mathrm{DIV}(\pi_{\theta})$ in Equation~\ref{eq:constrained_optimization} with its lower bound  $\mathbb{E}_{s\sim\rho^\pi(s|z),z\sim p(z)}\left[\min_{z'\neq z}\log{\frac{p(z|s)}{p(z'|s)+p(z|s)}}\right]$. The new optimization problem can be written as below:
\begin{eqnarray}
\max_{\pi_\theta} J(\theta),\ \mathrm{s.t.}\  \mathbb{E}_{s\sim\rho^{\pi}(s), z\sim p(z), a\sim\pi(\cdot|s,z)}[\sum_t\gamma^t r^{in}_t] \geq \delta.
\label{eq:new_constrained_optimization}
\end{eqnarray}

We can use a Lagrange multiplier $\lambda$ to solve the constrained optimization problem above:
\begin{eqnarray}
\begin{aligned}
& \max_{\pi_\theta}\min_{\lambda\geq0}  J(\theta) + \\
&\lambda(\mathbb{E}_{s\sim\rho^{\pi}(s), z\sim p(z), a\sim\pi(\cdot|s,z)}\left[\sum_t\gamma^t r^{in}_t\right]-\delta) \\
&= \max_{\pi_\theta}\min_{\lambda\geq0}  \mathbb{E}_{s\sim\rho^{\pi}(s),z\sim p(z), a\sim\pi(\cdot|s,z)}[ \\
&\sum_t\gamma^t r(s_t,a_t) + \lambda(\sum_t\gamma^t r^{in}_t-\delta)] \\
&\geq \max_{\pi_\theta}\mathbb{E}_{s\sim\rho^{\pi}(s),z\sim p(z)}[ \\
&\min_{\lambda\geq0} \mathbb{E}_{a\sim\pi(\cdot|s,z)}[\sum_t\gamma^t r(s_t,a_t)]+\lambda(\sum_t\gamma^t r^{in}_t-\delta)],
\label{eq:lagrange_constrained_optimization}
\end{aligned}
\end{eqnarray}
where the inequality on the last line is held because of Jensen's inequality. For the formula in the third line, it is difficult to estimate the average performance of the policy across all state spaces. However, in the formula of the fourth line, we have exchanged the order of the expectation and the $\min$, allowing us to compute based on each sampled data. Note that there are no similar issues in the external-reward-constraints optimization problem.

\section{F: Metrics}
\label{ap:Diversity metrics}
In this section, we first introduce the diversity metrics used in each environment.

{\bf MPE} $M_{Div}(\Pi) = \frac{1}{\mathit{n_z}}\sum_{i=1}^{\mathit{n_z}}\sum_{j=i+1}^{\mathit{n_z}}\ln(\Vert\Phi(\pi_i)-\Phi(\pi_j)\Vert_2)$, where $\Phi(\pi)$ is the behavior embedding of the policy $\pi$. In the MPE tasks, $\Phi(\pi)$ stands for an episode of agents' position collected by policy $\pi$. 

{\bf Atari} $M_{Div}(\Pi) = \frac{1}{\mathit{n_z}}\sum_{i=1}^{\mathit{n_z}}\sum_{j=i+1}^{\mathit{n_z}}(\Vert\Phi(\pi_i)-\Phi(\pi_j)\Vert_2)^2$, where $\Phi(\pi)$ is the behavior embedding of the policy $\pi$. In the Atari tasks, $\Phi(\pi)$ stands for an episode of the agents' center of mass collected by policy $\pi$. (In Pong, we collect the mass center of the green paddle in each step. In Boxing, we collect the mass center of the white boxer in each step)

{\bf SMAC} $M_{Div}(\Pi) = \frac{1}{\mathit{n_z}}\sum_{i=1}^{\mathit{n_z}}\sum_{j=i+1}^{\mathit{n_z}}(\Vert\Phi(\pi_i)-\Phi(\pi_j)\Vert_2)^2$, where $\Phi(\pi)$ is the behavior embedding of the policy $\pi$. In the SMAC tasks, $\Phi(\pi)$ stands for an episode of agents' position collected by policy $\pi$.

The chosen metric is preferable to directly using the discriminator's output as a diversity metric for two reasons. Firstly, deep neural networks can overfit and make the discriminator distinguish between strategies that appear similar in behavior. Consequently, the discriminator cannot accurately reflect the diversity score. Secondly, the metric is designed based on expert knowledge, meeting people's expectations of diverse strategies. It's important to note that the discriminator's input is independent of expert knowledge. In summary, the selected metric addresses the issues of overfitting and incorporates expert knowledge to ensure it captures the desired diversity in strategies.

\clearpage

\section{G: Algorithm}
\label{ap:algorithm}

\begin{algorithm}[h]
\caption{The DGPO Algorithm}
\label{algo: DGPO}
\begin{algorithmic}
\State{\bf Initialize:} Discriminator network parameters: $\phi$, actor network parameters: $\theta$, intrinsic critic network parameters: $\psi^{in}$ and extrinsic critic network parameters: $\psi^{ex}$.
\State{\bf Initialize:} replay memory $\mathcal{M}\leftarrow \{\}$.
\While{episode$\leq$ maximum episode}
    \State Sample a latent variable $z\sim p(z)$.
    \State Reset the environment and get states $s_t$.
    \While{$t\leq$ maximum timestep of roll-out}
        \State Sample action $a_t \sim \pi_{\theta}(a_t|s_{1:t},z)$ and apply to the environment.
        \State Receive $s_{t+1}$ and the extrinsic reward $r^{ex}_{t}$.
        \State Compute intrinsic reward $r^{in}_t$ from Eq.5.
        \State Store the transition 
        \State $\mathcal{M} \gets (s_t, a_t, r_t^{in}, r^{ex}_t, s_{t+1}, z)$.
    \EndWhile
    \While{$t\leq$ maximum timestep of training}
        \State Compute the total reward $r^{total}_t$ from Eq.15.
        \State Compute the expected return $R^{total}_{exp}, R^{ex}_{exp}, R_{exp}^{in}$ with $r^{total}_t, r^{ex}_t, r^{in}_t$ respectively. 
        \State Randomly sample a minibatch batch $\mathcal{B}_{data} \sim \mathcal{M}$, to train actor, critic and discriminator networks. 
        \State Compute PPO actor loss with $R^{total}_{exp}$ and critic losses with $R_{exp}^{in}$ and $R_{exp}^{ex}$.
        \State Update  $\theta$, $\psi^{in}$ and $\psi^{ex}$ 
        \State Compute discriminator loss via Eq.14.
        \State Update $\phi$ for the discriminator.
    \EndWhile
\EndWhile
\end{algorithmic}
\end{algorithm}
$R_{exp}^{total}, R_{exp}^{ex}, R_{exp}^{in}$ stands for the expected total return, expected external return, and expected internal return, respectively.

\section{H: Choice of Hyperparameter}
\label{ap:Hyperparameter}

Here are the guidelines for setting the hyperparameters $R_{target}$ and $\delta$. For $R_{target}$, we start by training an agent using MAPPO. Then, we calculate the expected value of the value functions. If the value is positive, we multiply it by 0.9, and if the value is negative, we multiply it by 1.1 to obtain the value of $R_{target}$. 

In the experiment, we found that the selection of $\delta$ was relatively difficult, and we found that setting the $\gamma$ of $J(\theta)$ in $mask_r$ to zero can make the process of selecting hyper-parameter simple and has no impact on performance.

At the beginning of training, the diversity scores $r_t^{in}$ are $\log(0.5)$ since the discriminator's distribution is uniform. As the training progresses, the value of $r_t^{in}$ converges to $\log(1)$. To determine the final value of $\delta$, we consider three candidate diversity thresholds: $\log(0.7)$, $\log(0.8)$, and $\log(0.9)$. We perform a parameter sweep on these choices to select the optimal value for $\delta$.

\section{I: Navigating $n_z=10$ Scenarios: Algorithmic Performance Evaluation}
\label{ap:nz10}

We tested the algorithm's performance in $n_z=10$ scenarios. The tests were conducted in two different environments: the $\text{Boxing}$ scene in $\text{Atari}$ and the $\text{Walker.Walk}$ scene in $\text{dm\_control}$.

\begin{figure}[h]
  \centering
  \includegraphics[width=0.98\linewidth]{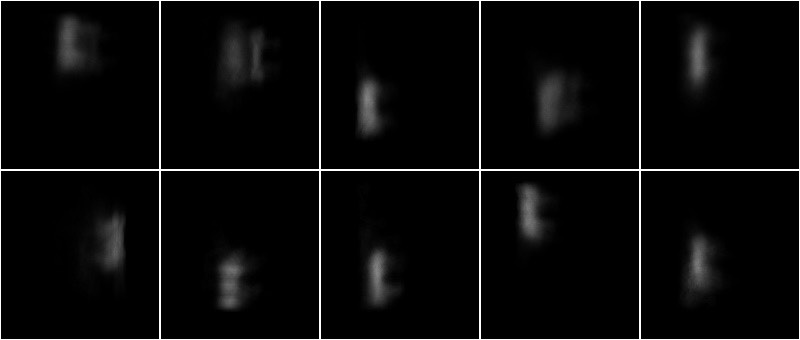}
  \vspace{-1mm}
  \caption{the algorithm's performance in the $\text{Boxing}$ scene}
  \label{fig:boxing_nz10}
  \vspace{-1mm}
\end{figure}

Figure~\ref{fig:boxing_nz10} illustrates the algorithm's performance in the $\text{Boxing}$ scene. The figure is divided into ten grids, each representing a diverse strategy. From left to right and top to bottom, the grids display strategies from z=0 to z=9. The white regions in the figure depict the average trajectories of the agent's movement. It's evident that across various $z$ values, the agent learns to attack the opponent from different directions. For this task, the average reward among the ten strategies (ten $z$ values) is $86.27$ points.

\begin{figure}[h]
  \centering
  \includegraphics[width=0.98\linewidth]{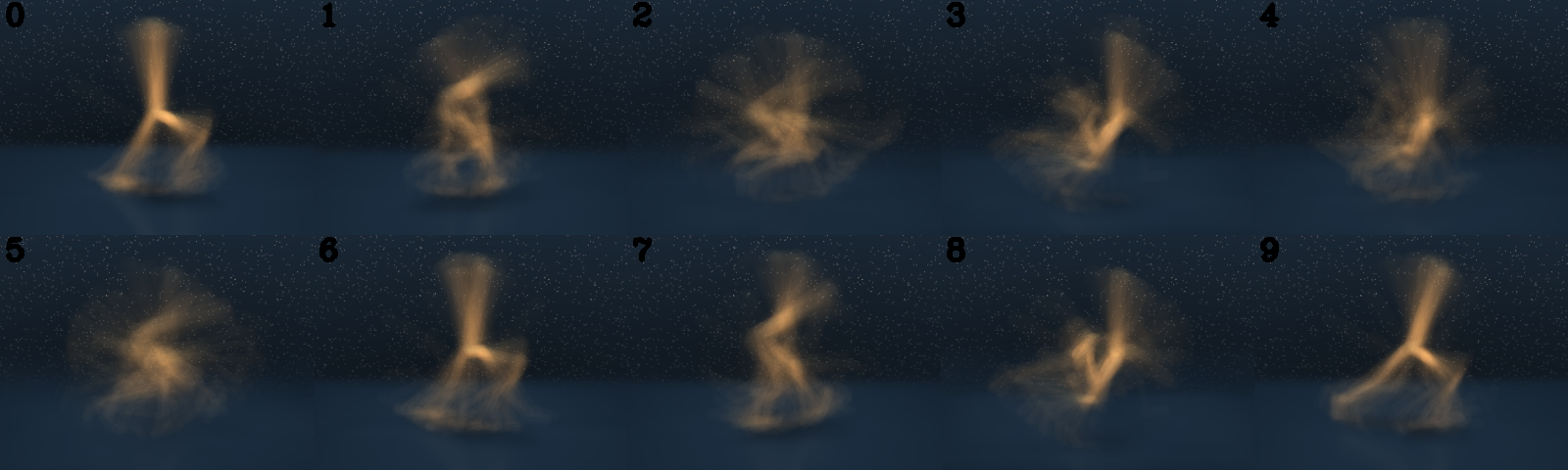}
  \vspace{-1mm}
  \caption{the algorithm's performance in the $\text{Walker.Walk}$ scene}
  \label{fig: dm_control}
  \vspace{-1mm}
\end{figure}

Figure~\ref{fig: dm_control} presents the algorithm's performance in the $\text{Walker.Walk}$ scene. Similar to Figure~\ref{fig:boxing_nz10}, there are ten grids, each representing a diverse strategy. The strategies range from $z=0$ to $z=9$, displayed from left to right and top to bottom. The figure showcases the average movement trajectories of the agent. It's noticeable that the agent adopts distinct movement patterns for different $z$ values. Conducting experiments in the MuJoCo environment also demonstrates the algorithm's capability to address tasks involving continuous action spaces. For this task, the average reward among the ten strategies (ten $z$ values) is $759.3$ points.

Additionally, we validated the ability to employ various strategies (different $z$ values) within the same trajectory during testing. This approach offers advantages such as adaptability in adversarial games, where we can dynamically shift our strategies over time, making it less predictable for opponents to exploit vulnerabilities. Specifically, we first trained ten distinct strategies using DGPO (with $n_z=10$ settings). Subsequently, during testing, we alternated the $z$ value at fixed intervals. The experimental outcomes are available in Appendix, $\text{video/boxing\_switch.avi}$. The results demonstrate that during testing, we can freely change $z$ values to acquire trajectories with enhanced diversity.

\end{document}